\documentclass[letterpaper]{article} 
\usepackage[preprint]{aaai2027}  
\usepackage[hyphens]{url}  
\usepackage{graphicx} 
\usepackage{natbib}  
\usepackage{caption} 
\usepackage{subcaption}
\usepackage{amsmath}
\usepackage{amssymb}
\usepackage{algorithm}
\usepackage{algorithmic}
\newcommand{\method}{CIPO}
\newcommand{\cmark}{\checkmark}
\usepackage{newfloat}
\usepackage{listings}
\DeclareCaptionStyle{ruled}{labelfont=normalfont,labelsep=colon,strut=off} 
\floatstyle{ruled}
\newfloat{listing}{tb}{lst}{}
\floatname{listing}{Listing}

\usepackage{booktabs}
\usepackage[table]{xcolor}
\usepackage{multirow}  
\usepackage{arydshln}  
\title{Contextual Information Policy Optimization for Search Agents}
\author {
    Xingyu Guo\textsuperscript{\rm 1},
    Wei Chen\textsuperscript{\rm 2},
    Linlin Yang\textsuperscript{\rm 3}\corresponding,
    Baochang Zhang\textsuperscript{\rm 2}
}
\affiliations {
    \textsuperscript{\rm 1}National College for Excellent Engineers, Beihang University,
Beijing, China\\
    \textsuperscript{\rm 2}School of Artificial Intelligence, Beihang University, Beijing, China \\
    \textsuperscript{\rm 3}State Key Laboratory of Media Convergence
and Communication, Communication University of China, Beijing, China \\
    22373140@buaa.edu.cn
}
\begin{document}

\maketitle

\begin{abstract}
    Search agents extend large language models beyond static parametric memory by enabling them to acquire and use external evidence during multi-step reasoning. For knowledge-intensive tasks involving complex or evolving information, their reliability depends not only on retrieving relevant evidence but also on using it to guide subsequent reasoning. However, existing methods primarily reward final-answer correctness or intermediate progress, without directly assessing whether post-retrieval actions are grounded in the retrieved evidence. This misalignment encourages prior-driven reasoning: agents form conclusions based on internal knowledge and use retrieval mainly to confirm them, resulting in confirmation bias and inefficient evidence use.
To address this issue, we propose Contextual Information Policy Optimization (CIPO), an evidence-oriented reinforcement learning framework that explicitly aligns policy optimization with external evidence use. CIPO assigns dense, turn-level credit to reasoning actions influenced by retrieved information, while combining this evidence-use signal with a global outcome reward to preserve answer correctness. With this manner, CIPO discourages evidence-detached guesses and promotes reasoning trajectories in which retrieved facts can guide or revise subsequent reasoning. Importantly, CIPO requires neither human process annotations nor an additional reward model. Extensive experiments on seven in-domain and out-of-domain benchmarks show that CIPO reduces the prevalence of prior-driven reasoning and achieves excellent performance on most tasks. 

\begin{links}
    \link{Code}{https://github.com/gxingyu/cipo}
\end{links}

\end{abstract}


\section{Introduction}

Large language models (LLMs) are increasingly equipped with external retrieval tools \citep{lewis2020retrieval}, transforming them from passive text generators into search agents that actively interact with external knowledge sources during reasoning \citep{yao2022react,trivedi2023interleaving}. For tasks requiring complex or up-to-date knowledge, these agents can mitigate the limitations of static parametric memory by dynamically determining when and what to retrieve, extracting useful evidence from noisy retrieval results, and incorporating external information into their reasoning trajectories \citep{mallen2023popqa,asai2024selfrag}. However, effective search requires agents to coordinate retrieval and reasoning through a complex, dynamic sequence of decisions. Reinforcement learning \citep{jin2025searchr1} provides a natural paradigm for training search agents, as it enables them to explore this decision space through trial and error, and learn how to interleave tool use with multi-step reasoning \citep{wang2026igpo, he2026searchr2}. In this way, search agents can progressively develop the ability to solve knowledge-intensive tasks with reduced human intervention.

\begin{figure}[!t]
	\centering
	\includegraphics[width=\columnwidth]{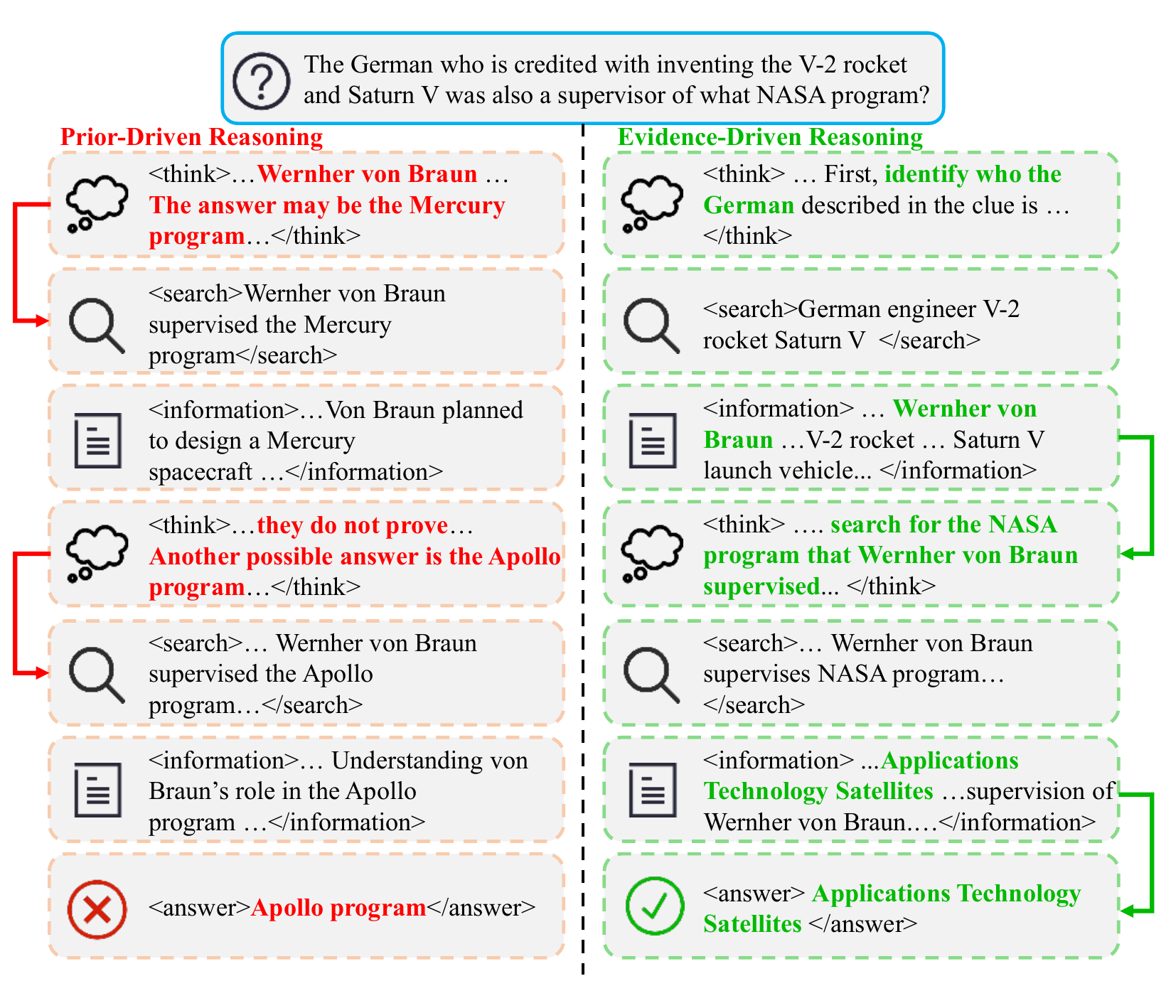}
	\caption{Illustrative case for the same question. Prior-driven reasoning uses retrieval mainly to confirm a parametric guess, whereas evidence-driven reasoning lets retrieved evidence change the next action.}
	\label{fig:evidence_case}
\end{figure}

Existing reinforcement learning approaches for search agents broadly fall into two paradigms. The first relies primarily on sparse outcome rewards \cite{jin2025searchr1,zheng2025deepresearcher}, and optimizes multi-turn interaction trajectories by evaluating the correctness of the final answer. However, the credit assignment problem in long-context reasoning makes supervision based solely on the final outcome excessively sparse. To address this limitation, the second paradigm \cite{feng2026gigpo,wang2026igpo} introduces denser supervision at reasoning steps in addition to outcome rewards. These methods seek to assign process rewards to intermediate reasoning steps that move the trajectory toward the correct answer. Although both paradigms can improve agent performance, they share a fundamental limitation: prior driven reasoning caused by reward misalignment. Specifically, existing objectives assess only final answer correctness or the probability that a trajectory converges toward the correct answer. They do not directly evaluate whether reasoning actions after retrieval genuinely depend on newly retrieved evidence. Consequently, an agent can receive high rewards even when it forms a hypothesis solely from internal parametric knowledge and reduces subsequent retrieval to a mere confirmation procedure for that generated hypothesis.

This reward misalignment substantially degrades the use of external knowledge and predisposes agents to confirmation bias. A model can reach a conclusion entirely from its internal prior knowledge, while retrieval merely endorses that conclusion and fails to correct or guide subsequent reasoning claims. As illustrated in Figure~\ref{fig:evidence_case}, prior driven reasoning differs fundamentally from the desired evidence driven reasoning. In a prior driven trajectory, the model appears to invoke a search, and relevant supporting facts appear in the context, but its actual reasoning does not incorporate this external information and merely confirms its initial hypothesis. When the model has uncertain parametric memory or hallucinated knowledge about facts in the long tail, such seemingly valid but spurious reasoning is harmful and leaves high-quality retrieved evidence entirely unused \citep{mallen2023popqa}. Continuing to assign full rewards to such misaligned trajectories further reinforces confirmation bias. This limitation creates an urgent need for a new training mechanism that penalizes reasoning that ignores newly retrieved evidence and assigns appropriate credit to steps that genuinely use it.

To this end, we propose Contextual Information Policy Optimization (CIPO), an evidence-oriented reinforcement learning framework that rewards search agents for substantively using retrieved external evidence at intermediate reasoning steps. For each reasoning step after retrieval, CIPO constructs a local contrastive signal by comparing the generation likelihood of an action when evidence is visible with that when evidence is masked. This comparison yields the Evidence-Access Log-Likelihood Ratio (EALR), which directly quantifies the sensitivity of the current reasoning action to the most recently retrieved passages without requiring an external reward model or human annotations. CIPO converts EALR into dense rewards at each reasoning turn and combines these rewards with global outcome supervision for the final answer. It therefore penalizes ineffective reasoning steps that make guesses from internal parametric knowledge without actually using retrieved external evidence.


In summary, the main contributions of this paper are as follows:
\textbf{(i)} We identify reward misalignment caused by prior-driven reasoning and propose CIPO, an evidence-oriented reinforcement learning framework that promotes the use of retrieved evidence.
\textbf{(ii)} We develop EALR, a dense step-level reward that compares action likelihoods with and without retrieved evidence, requiring neither human annotations nor additional rollouts.
\textbf{(iii)} Experiments on seven in-domain and out-of-domain benchmarks show that CIPO substantially reduces prior-driven reasoning and outperforms state-of-the-art baselines on most tasks.

\begin{figure*}[!t]
	\centering
	\includegraphics[page=1,width=0.99\textwidth]{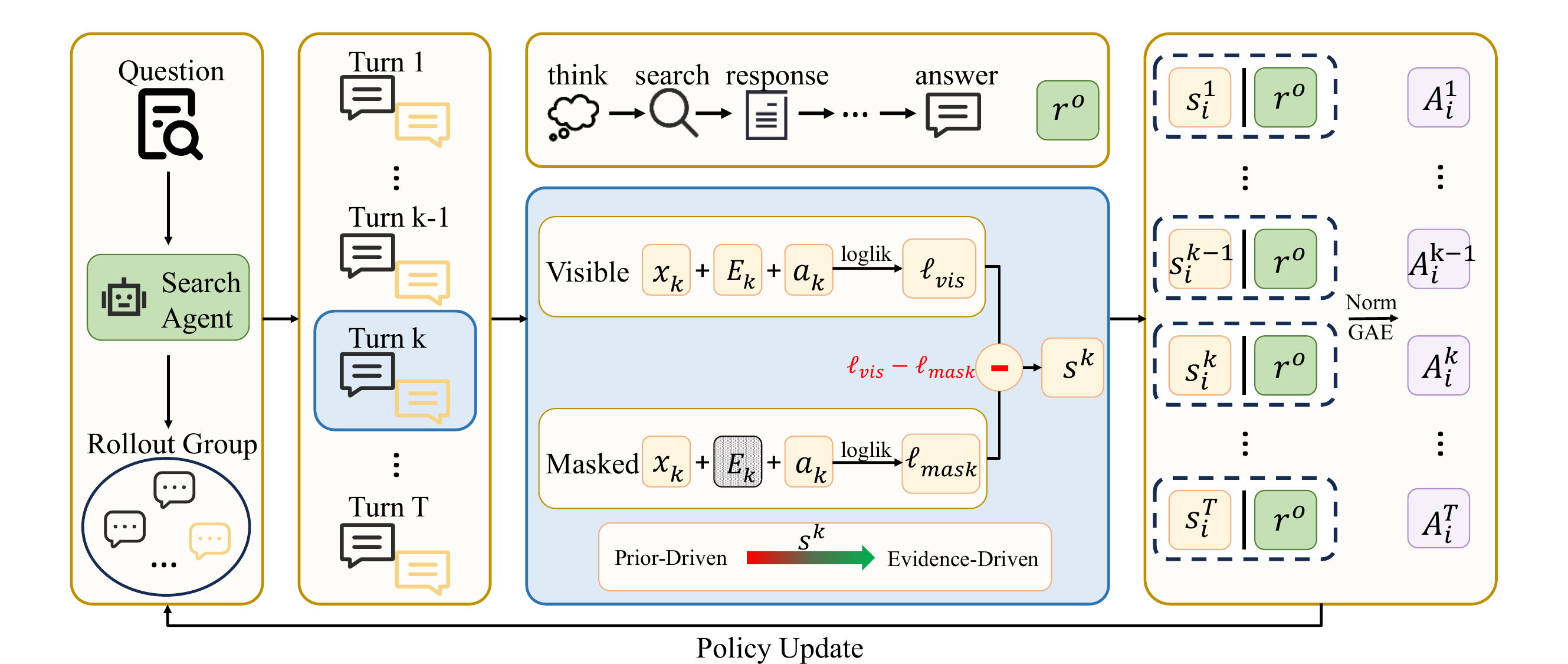}
	\caption{The overall architecture of CIPO. It evaluates each post-retrieval reasoning step under evidence-visible and evidence-masked conditions, computes an EALR-based reward, and integrates it with outcome supervision for policy optimization.}
	\label{fig:main_architecture}
\end{figure*}

\section{Related Work}

\paragraph{Search-Integrated Reasoning.}
RAG systems condition language-model outputs on passages retrieved from external corpora \citep{lewis2020retrieval}, with subsequent studies surveying its broader design space and empirical best practices \citep{gao2023ragsurvey,wang2024ragbestpractices}. ReAct \citep{yao2022react}, IRCoT \citep{trivedi2023interleaving}, and Self-RAG \citep{asai2024selfrag} further integrate retrieval with iterative reasoning, while EviNote-RAG \citep{dai2025evinoterag} and CoRAG \citep{wang2024corag} improve evidence organization and retrieval efficiency. More recent RL-based agents, including Search-R1 \citep{jin2025searchr1},
R1-Searcher \citep{song2025r1searcher}, DeepResearcher
\citep{zheng2025deepresearcher}, and Search-R2 \citep{he2026searchr2},
learn to formulate queries and interact with external search environments.
Recent work further jointly optimizes retrievers and reasoners or introduces
retrieval-aware intermediate supervision
\citep{zeng2026cosearch,goel2026s,wang2026pathrouter}.
This direction has expanded to web, tool-use, and deep-research settings
\citep{qi2025webrl,li2025torl,dai2025carefulqueries,deng2025atomsearcher,
zhang2025agenticdeepresearch}, with recent surveys providing broader
taxonomies of these agents
\citep{huang2025deepresearchsurvey,ning2025webagentssurvey,
qu2025toolsurvey,zheng2024openresearcher}. Despite this progress, these
methods primarily optimize retrieval behavior, answer accuracy, or
retrieval-derived credit signals, without directly measuring whether a
specific retrieved observation changes the distribution of the agent's
subsequent reasoning action.

\paragraph{Credit Assignment in Multi-Turn RL.}
LLM agents are commonly optimized with PPO \citep{schulman2017ppo}, GRPO \citep{shao2024deepseekmath} and RLOO \citep{ahmadian2024rloo}, while systems such as DeepSeek-R1 \citep{guo2025deepseekr1} and DAPO \citep{yu2026dapo} demonstrate the scalability of reinforcement learning with verifiable rewards. However, terminal rewards provide limited guidance for long interaction trajectories, motivating process supervision in mathematical reasoning and agentic RAG \citep{wang2023mathshepherd,zhang2026process}. Search-specific methods provide denser feedback through intermediate rewards or finer-grained advantage estimation: StepSearch \citep{wang2025stepsearch} introduces step-wise optimization, ReasonRAG \citep{zhang2026process} supervises multiple stages of retrieval and generation, GiGPO \citep{feng2026gigpo} estimates group-relative step advantages, and IGPO \citep{wang2026igpo} measures turn-wise progress toward the correct answer. Related analyses connect intermediate probability gains with reasoning progress \citep{gan2025rethinking}, and broader surveys summarize emerging agentic RL objectives \citep{zhang2025agenticrlsurvey}. Yet, these signals primarily measure task progress or correctness rather than the dependence of reasoning actions on newly retrieved evidence available to the agent at each reasoning step.

\section{Proposed Method}

In this section, we present \method{}, an evidence-oriented reinforcement
learning framework for search agents. The key idea is to distinguish actions
that genuinely depend on retrieved evidence from those driven primarily by
the model's parametric prior. \method{} first quantifies the
effect of retrieved information on subsequent actions through EALR, and then
integrates this turn-level signal with outcome supervision for policy
optimization, as illustrated in Figure~\ref{fig:main_architecture}.

\subsection{EALR Reward}

To measure whether retrieved information affects subsequent decisions, we
first represent each trajectory $\tau_i$ as a sequence of $T_i$ interaction
turns. This turn-level formulation allows us to compare the model's
post-retrieval reasoning and search actions under evidence-visible and
evidence-masked conditions, with all preceding trajectory context held fixed:
\begin{equation}
	\mathcal{R}_{i,t}
	=
	B^{\mathrm{think}}_{i,t}
	\oplus
	B^{\mathrm{search}}_{i,t}
	\oplus
	E_{i,t},
	\qquad
	t=1,\ldots,T_i,
\end{equation}
where the three spans correspond to the tagged \texttt{<think>}, \texttt{<search>}, and \texttt{<information>} blocks, respectively, and $\oplus$ denotes sequence concatenation.

For each $t<T_i$, let $h_{i,t}$ denote the complete trajectory prefix ending at the \texttt{<search>} block of turn $t$. Thus, $h_{i,t}$ contains the original query, all preceding turns, and the current \texttt{<think>} and \texttt{<search>} blocks, but not the current information block $E_{i,t}$.

We evaluate whether $E_{i,t}$ affects the policy-generated blocks in the next turn. We further analyze the choice of next-turn action as the EALR target and compare it with full-trajectory scoring in Appendix D. Specifically, the target action is
\begin{equation}
	a_{i,t+1}
	=
	B^{\mathrm{think}}_{i,t+1}
	\oplus
	B^{\mathrm{search}}_{i,t+1}
	=
	(y_1,\ldots,y_m).
\end{equation}
The target includes all tokens in the next \texttt{<think>} and \texttt{<search>} blocks. Environment-generated \texttt{<information>} tokens are never included in the scored action.

We score the same observed action under two otherwise identical conditions. In the evidence-visible condition, the action has normal causal access to the information block from the preceding turn:
\begin{equation}
	\ell_{\mathrm{vis}}(a_{i,t+1})
	=
	\sum_{j=1}^{m}
	\log
	\pi_{\theta_{\mathrm{old}}}
	\left(
	y_j
	\mid
	h_{i,t},E_{i,t},y_{<j}
	\right).
\end{equation}

In the evidence-masked condition, we retain the original token sequence and
apply an additive attention mask during teacher-forced scoring. At every
Transformer layer and attention head, each token in the target action
$a_{i,t+1}$ is prevented from attending to any token in the preceding
information block $E_{i,t}$ by setting the corresponding attention logits to
$-\infty$ before the softmax operation. All other attention relations remain
unchanged: target tokens retain access to the trajectory prefix $h_{i,t}$ and
to preceding target tokens $y_{<j}$.

We denote this attention-masked condition by $M(E_{i,t})$. Its action log-likelihood is
\begin{equation}
	\ell_{\mathrm{mask}}(a_{i,t+1})
	=
	\sum_{j=1}^{m}
	\log
	\pi_{\theta_{\mathrm{old}}}
	\left(
	y_j
	\mid
	h_{i,t},M(E_{i,t}),y_{<j}
	\right).
\end{equation}

Masking does not delete or replace the information tokens. Token identities, sequence length, position indices, preceding trajectory history, and model parameters remain unchanged. Only the next-turn action tokens' attention access to the immediately preceding \texttt{<information>} block is removed.

We define the token-averaged EALR for the transition from turn $t$ to turn $t+1$ as
\begin{equation}
	\begin{aligned}
		s^{\mathrm{ealr}}_{i,t}
		&=
		\frac{1}{m}
		\left[
		\ell_{\mathrm{vis}}(a_{i,t+1})
		-
		\ell_{\mathrm{mask}}(a_{i,t+1})
		\right] \\
		&=
		\frac{1}{m}
		\log
		\frac{
			\pi_{\theta_{\mathrm{old}}}
			(a_{i,t+1}\mid h_{i,t},E_{i,t})
		}{
			\pi_{\theta_{\mathrm{old}}}
			(a_{i,t+1}\mid h_{i,t},M(E_{i,t}))
		}.
	\end{aligned}
\end{equation}

Both likelihoods are computed by teacher forcing the same collected next-turn action through the frozen rollout policy $\pi_{\theta_{\mathrm{old}}}$. Therefore, EALR requires neither an additional trajectory rollout nor an external reward model. The resulting reward is detached and kept fixed during the corresponding policy update to maintain a fixed credit signal.

A positive EALR means that access to the retrieved information increases the likelihood of the observed next-turn reasoning and search action. A value near zero indicates that the same action is almost equally likely without access to that information, while a negative value indicates that masking the information makes the action more likely.

For the sequence-level log-likelihood ratio,
\begin{equation}
	S(a;h,E)
	=
	\log
	\frac{
		\pi_{\theta_{\mathrm{old}}}(a\mid h,E)
	}{
		\pi_{\theta_{\mathrm{old}}}(a\mid h,M(E))
	}.
\end{equation}

Its expectation under the evidence-visible policy is
\begin{equation}
	\begin{aligned}
		\mathbb{E}_{
			a\sim\pi_{\theta_{\mathrm{old}}}
		}
		\left[
			S(a;h,E)
		\right]
		&=
		D_{\mathrm{KL}}
		\Bigl(
			\pi_{\theta_{\mathrm{old}}}(\cdot\mid h,E)
			\,\Vert \\
		&\qquad
			\pi_{\theta_{\mathrm{old}}}(\cdot\mid h,M(E))
		\Bigr).
	\end{aligned}
\end{equation}

EALR is a length-normalized policy-level evidence-access signal. It
quantifies the likelihood contrast between evidence-visible and
evidence-masked conditions for the same post-retrieval reasoning and search action.

\subsection{Turn-Level Advantage Allocation}

Having obtained an EALR score for each post-retrieval transition, we next
combine this evidence-use signal with the terminal outcome reward. This design
provides dense turn-level feedback while retaining supervision for final-answer
correctness. To ensure that neither reward type dominates because of scale
differences, we normalize them separately within each rollout group before
constructing advantages.

For each query $q$, we sample a group of $G$ trajectories. Each trajectory
$i$ yields a terminal outcome reward $r_i^{out}$ at turn $T_i$ and EALR
scores $\{s^{\mathrm{ealr}}_{i,k}\}_{k<T_i}$ at post-retrieval turns. We use
$r_{i,k}^{\mathrm{ealr}}=s^{\mathrm{ealr}}_{i,k}$ as the step-level reward.
The EALR reward is detached and remains fixed during the corresponding policy
update.

Since $r^{\mathrm{ealr}}$ and $r^{out}$ differ in scale, normalizing them independently within the rollout group $G$ ensures stable relative advantage estimates. The normalized reward $\tilde{r}_{i,k}$ for the $k$-th turn is computed by standardizing the respective rewards across all valid turns in the group:
\begin{equation}
	\tilde{r}_{i,k} = 
	\begin{cases} 
		\frac{r_{i,k}^{\mathrm{ealr}} - \text{mean}(r^{\mathrm{ealr}})}{\text{std}(r^{\mathrm{ealr}})}, & 1 \le k < T_i, \\
		\frac{r_i^{out} - \text{mean}(r^{out})}{\text{std}(r^{out})}, & k = T_i.
	\end{cases}
\end{equation}

While $\tilde{r}_{i,k}$ captures the relative quality of each individual turn, it ignores the impact of current decisions on future turns. To incorporate such long-horizon dependencies, we compute the turn-level discounted return (advantage) $\hat{A}_{i,t}$ for turn $t$ via discounted accumulation:
\begin{equation}
	\hat{A}_{i,t} = \sum_{j=t}^{T_i} \gamma^{j-t} \tilde{r}_{i,j},
\end{equation}
where $\gamma \in (0, 1]$ is the discount factor. This formulation integrates local evidence-dependence feedback with global
task-success supervision, yielding turn-level advantages that propagate both
signals across the full reasoning trajectory.

\subsection{Policy Optimization Objective}
Building on the turn-level advantages defined above, CIPO optimizes the policy
using a clipped objective. Because retrieved evidence is returned by the environment rather than
generated by the policy, its tokens serve only as contextual observations and
are excluded from the optimization loss. We apply each turn-level advantage to
the model-generated reasoning and search tokens within the corresponding turn
of the collected trajectory.

We broadcast the advantage $\hat{A}_{i,t}$ to all model-generated decision tokens within turn $t$. Let $\mathcal{M}_i$ denote the set of model-generated token indices in trajectory $i$, and let $t(s)$ be a mapping function from token index $s$ to its corresponding turn. For each decision token $y_{i,s}$, the importance sampling ratio is defined as $\rho_{i,s}(\theta) = \frac{\pi_\theta(y_{i,s}|h_{i,s})}{\pi_{\theta_{old}}(y_{i,s}|h_{i,s})}$, where $h_{i,s}$ denotes the historical context encompassing the query and all preceding tokens up to step $s$ in trajectory $i$. Here, $\theta$ and $\theta_{old}$ represent the parameters of the current policy and the old behavior policy used for trajectory sampling, respectively. We first define the clipped loss for a single decision token $s$:
\begin{equation}
	\begin{aligned}
		\mathcal{L}^{CLIP}_{i,s}(\theta) &= \min \Big( \rho_{i,s}(\theta) \hat{A}_{i,t(s)}, \\
		&\quad \text{clip}(\rho_{i,s}(\theta), 1-\epsilon, 1+\epsilon) \hat{A}_{i,t(s)} \Big),
	\end{aligned}
\end{equation}
where $\epsilon$ is a hyperparameter determining the clipping bound to prevent destructively large policy updates.

The overall CIPO objective is then defined as the expected return minus a KL divergence penalty:
\begin{equation}
	\begin{aligned}
		\mathcal{J}_{CIPO}(\theta) &= \mathbb{E}_{q, \tau} \Bigg[ \frac{1}{G} \sum_{i=1}^G \frac{1}{|\mathcal{M}_i|} \sum_{s \in \mathcal{M}_i} \mathcal{L}^{CLIP}_{i,s}(\theta) \\
		&\quad - \beta \mathbb{D}_{KL}(\pi_\theta || \pi_{ref}) \Bigg],
	\end{aligned}
\end{equation}
where $\mathbb{E}_{q, \tau}$ is shorthand for $\mathbb{E}_{q \sim \mathcal{D}, \tau \sim \pi_{\theta_{old}}}$, indicating the expectation over queries $q$ from the dataset $\mathcal{D}$ and trajectories $\tau$ sampled from the old policy. $G$ is the group size of rollouts per query. $\beta$ is the coefficient for the per-token Kullback-Leibler (KL) divergence penalty against the reference model $\pi_{ref}$, which prevents policy over-optimization and preserves generation fluency. Optimizing $\mathcal{J}_{CIPO}$ combines task-success supervision with relative credit for actions that are more sensitive to accessible external evidence.

\begin{table*}[!t]
	\centering
	\normalsize
	\setlength{\tabcolsep}{8.0pt}
	\renewcommand{\arraystretch}{0.95}
		\begin{tabular}{lcccccccc}
			\toprule
			\multirow{2}{*}{Methods}
			& \multicolumn{4}{c}{In-domain}
			& \multicolumn{3}{c}{Out-of-domain}
			& \multirow{2}{*}{Avg.} \\
			\cmidrule(lr){2-5} \cmidrule(lr){6-8}
			& NQ & TQ & HotpotQA & 2Wiki & MuSiQue & Bamboogle & PopQA & \\
			\midrule
			\multicolumn{9}{l}{\textbf{Qwen2.5-7B-Base/Instruct}} \\
			CoT & 0.252 & 0.301 & 0.284 & 0.261 & 0.155 & 0.354 & 0.372 & 0.283 \\
			CoT+RAG & 0.312 & 0.385 & 0.364 & 0.342 & 0.201 & 0.402 & 0.461 & 0.352 \\
			Search-R1-base & 0.331 & 0.447 & 0.457 & 0.434 & 0.265 & 0.450 & 0.430 & 0.402 \\
			Search-R1-instruct & 0.285 & 0.362 & 0.401 & 0.382 & 0.223 & 0.481 & 0.472 & 0.372 \\
			R1-searcher & 0.345 & 0.461 & 0.462 & 0.441 & 0.281 & 0.472 & 0.456 & 0.417 \\
			DeepResearcher & 0.362 & 0.483 & 0.471 & 0.452 & 0.274 & 0.488 & 0.455 & 0.426 \\
			StepSearch-base & 0.381 & 0.521 & 0.482 & \textbf{0.461} & 0.291 & 0.485 & 0.438 & 0.437 \\
			StepSearch-instruct & 0.321 & 0.422 & 0.425 & 0.401 & 0.252 & 0.463 & 0.428 & 0.387 \\
			ReasonRAG & 0.401 & 0.552 & \underline{0.493} & 0.421 & \underline{0.302} & 0.490 & 0.465 & 0.446 \\
			GiGPO & 0.423 & 0.601 & 0.491 & 0.412 & 0.285 & 0.495 & 0.461 & 0.453 \\
			IGPO & \underline{0.447} & \underline{0.683} & 0.482 & 0.385 & 0.213 & \underline{0.501} & \textbf{0.488} & \underline{0.457} \\
			\hdashline
			 \textbf{\method{}(Ours)} & \textbf{0.467} & \textbf{0.726} & \textbf{0.521} & \underline{0.460} & \textbf{0.322} & \textbf{0.552} & \underline{0.482} & \textbf{0.504} \\
			\specialrule{\lightrulewidth}{0pt}{\belowrulesep}
			\multicolumn{9}{l}{\textbf{Qwen2.5-3B-Base/Instruct}} \\
			CoT & 0.247 & 0.270 & 0.239 & 0.248 & 0.148 & 0.274 & 0.364 & 0.256 \\
			CoT+RAG & 0.306 & 0.346 & 0.306 & 0.325 & 0.192 & 0.311 & 0.450 & 0.319 \\
			Search-R1-base & 0.325 & 0.401 & 0.384 & 0.412 & 0.253 & 0.348 & 0.420 & 0.363 \\
			Search-R1-instruct & 0.279 & 0.325 & 0.337 & 0.363 & 0.213 & 0.372 & \underline{0.461} & 0.336 \\
			R1-searcher & 0.338 & 0.414 & 0.388 & 0.419 & 0.268 & 0.365 & 0.446 & 0.377 \\
			DeepResearcher & 0.355 & 0.434 & 0.396 & 0.429 & 0.261 & 0.380 & 0.445 & 0.386 \\
			StepSearch-base & 0.374 & 0.468 & 0.405 & \textbf{0.440} & 0.277 & 0.375 & 0.428 & 0.395 \\
			StepSearch-instruct & 0.315 & 0.379 & 0.357 & 0.381 & 0.240 & 0.358 & 0.418 & 0.350 \\
			ReasonRAG & 0.393 & 0.496 & \underline{0.414} & 0.400 & \underline{0.288} & 0.381 & 0.454 & 0.404 \\
			GiGPO & 0.415 & 0.540 & 0.413 & 0.391 & 0.272 & 0.383 & 0.450 & \underline{0.409} \\
			IGPO & \underline{0.438} & \underline{0.613} & 0.405 & 0.366 & 0.203 & \underline{0.388} & 0.387 & 0.400 \\
			\hdashline
		 \textbf{\method{}(Ours)} & \textbf{0.458} & \textbf{0.652} & \textbf{0.438} & \underline{0.437} & \textbf{0.307} & \textbf{0.427} & \textbf{0.471} & \textbf{0.456} \\
			\specialrule{\heavyrulewidth}{0pt}{0pt}
		\end{tabular}
	\caption{Main results on in-domain and out-of-domain question answering benchmarks. We report F1 scores. Bold numbers indicate the best result within each backbone block, and underlined numbers indicate the second-best result.}
	\label{tab:main_results}
\end{table*}

\section{Experiments}
\subsection{Experimental Setup}
\paragraph{Datasets \& Metrics.}
We evaluate the effectiveness of the proposed framework across both in-domain and out-of-domain question-answering benchmarks within an agentic search setting. The in-domain evaluation comprises Natural Questions \citep{kwiatkowski2019natural}, TriviaQA \citep{joshi2017triviaqa}, HotpotQA \citep{yang2018hotpotqa}, and 2WikiMultiHopQA \citep{ho2020constructing}. For the out-of-domain evaluation, we utilize MuSiQue \citep{trivedi2022musique}, Bamboogle \citep{press2023bam}, and PopQA \citep{mallen2023popqa}. We train on the
training splits of the in-domain datasets and evaluate on the test splits of
all seven datasets. To ensure consistency with prior work \citep{wang2026igpo},  We report the word-level F1 score as the primary evaluation metric, which is computed as the harmonic mean of precision and recall between the predicted responses and the reference answers.

\paragraph{Baselines.}
We compare \method{} with three categories of baselines. The first category
contains methods without search-specific RL training: chain-of-thought (CoT)
\citep{wei2022cot} and CoT augmented with retrieved evidence (CoT+RAG)
\citep{lewis2020retrieval}. The second category contains outcome-only RL search
agents, including Search-R1 \citep{jin2025searchr1}, R1-Searcher
\citep{song2025r1searcher}, and DeepResearcher
\citep{zheng2025deepresearcher}, which optimize search trajectories using
final-answer rewards. The third category contains process-supervised RL
methods that combine outcome rewards with intermediate supervision, including
StepSearch \citep{wang2025stepsearch}, ReasonRAG
\citep{zhang2026process}, GiGPO \citep{feng2026gigpo}, and IGPO
\citep{wang2026igpo}.

\paragraph{Implementation Details.}
We evaluate all methods with Qwen2.5-3B-Instruct and Qwen2.5-7B-Instruct
\citep{qwen2025qwen25} as backbone models. Training is conducted with the
verl framework \citep{sheng2025hybridflow} on 8 × NVIDIA A100-80G GPUs, using a discount factor of $\gamma=1.0$. At each training step, we sample 32 prompts and generate 16
rollouts per prompt, with at most five interaction turns per rollout. We use a local retrieval service consistent with the Search-R1 environment
architecture to provide reproducible search interactions. All implemented RL
methods use the same training data, retrieval environment, rollout-group
size, maximum number of turns, and training budget. Detailed optimization settings and prompt templates for training and inference
are provided in Appendix C for full reproducibility.

\begin{figure*}[!t]
	\centering
	
	\begin{subfigure}[t]{0.242\textwidth}
		\centering
		\includegraphics[width=\linewidth]
		{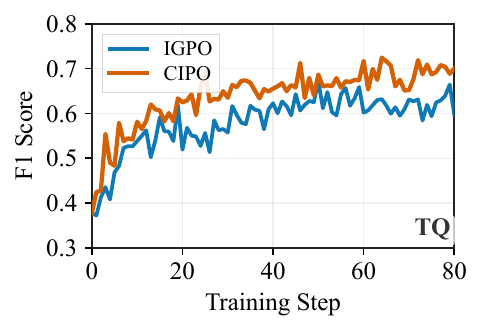}
		\caption{TriviaQA (ID)}
		\label{fig:f1_tq}
	\end{subfigure}\hfill
	\begin{subfigure}[t]{0.242\textwidth}
		\centering
		\includegraphics[width=\linewidth]
		{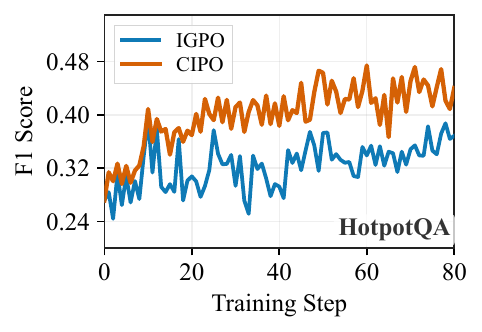}
		\caption{HotpotQA (ID)}
		\label{fig:f1_hotpotqa}
	\end{subfigure}\hfill
	\begin{subfigure}[t]{0.242\textwidth}
		\centering
		\includegraphics[width=\linewidth]
		{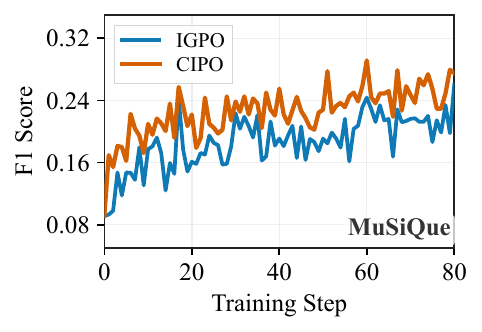}
		\caption{MuSiQue (OOD)}
		\label{fig:f1_musique}
	\end{subfigure}\hfill
	\begin{subfigure}[t]{0.242\textwidth}
		\centering
		\includegraphics[width=\linewidth]
		{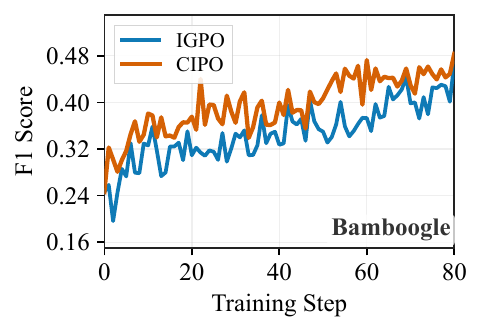}
		\caption{Bamboogle (OOD)}
		\label{fig:f1_bamboogle}
	\end{subfigure}
	
	\caption{Representative step-wise validation-set F1 trajectories
		on two in-domain and two out-of-domain benchmarks. Each panel
		compares \method{} with IGPO over 80 training steps. The vertical
		scale is selected separately for each dataset.}
	\label{fig:training_dynamics_main}
\end{figure*}

\subsection{Performance Comparison}

Table~\ref{tab:main_results} reports word-level F1 on four in-domain and
three held-out out-of-domain question-answering benchmarks. CIPO consistently achieves the highest macro-average F1 scores for both evaluated model sizes. Specifically, utilizing the Qwen2.5-7B-Instruct model allows CIPO to reach a score of 0.504, which exceeds the next best method (IGPO at 0.457) by 4.7 F1 points. Similarly, the Qwen2.5-3B-Instruct backbone enables CIPO to score 0.456, surpassing its strongest competitor (GiGPO at 0.409) by an identical margin of 4.7 points. This consistent improvement demonstrates that CIPO's advantages are effective
across model scales rather than being exclusive to larger architectures within
the evaluated Qwen2.5 model family.

When analyzing the individual datasets, the 7B iteration of CIPO secures the highest results on Natural Questions, TriviaQA, HotpotQA, MuSiQue, and Bamboogle. The 3B iteration replicates this success across these five benchmarks and additionally claims the top score on PopQA. Although StepSearch performs marginally better on 2WikiMultiHopQA for both
model scales, and IGPO achieves the top 7B result on PopQA, CIPO still
maintains the most substantial overall performance gains across the evaluated
benchmarks under both model configurations.

Importantly, CIPO demonstrates strong generalization capabilities in held-out, out-of-domain evaluations. Even though the training process relies entirely on in-domain datasets, the 7B model outperforms competitors on MuSiQue and Bamboogle, and the 3B model leads across all three out-of-domain benchmarks, including PopQA. These robust out-of-domain results validate the effectiveness of employing
EALR-based turn-level credit to improve the performance of search agents under
the evaluated retrieval setting.

\subsection{Ablation Study}

Table 2 evaluates the individual and combined effects of terminal outcome supervision and EALR. We additionally report evidence utilization metrics, including answer-supportive evidence utilization (Sup.) and irrelevant evidence utilization (Irr.), their definitions and evaluation protocol are provided in Appendix E. Compared with outcome-only training, EALR-only
training improves the average F1 from 0.336 to 0.393 for the 3B model and
from 0.372 to 0.430 for the 7B model. It also substantially increases
supportive-evidence utilization, from 18.6\% to 43.8\% for the 3B model and
from 21.9\% to 48.5\% for the 7B model. However, the corresponding
irrelevant-evidence utilization rates increase to 31.7\% and 28.9\%,
respectively, indicating that EALR alone encourages evidence responsiveness
but does not fully distinguish useful evidence from irrelevant content.

Combining EALR with outcome supervision yields the strongest performance
across all reported metrics. Relative to EALR-only training, the combined
objective improves the average F1 from 0.393 to 0.456 for the 3B model and
from 0.430 to 0.504 for the 7B model. More importantly, it further increases
supportive-evidence utilization to 55.2\% and 60.7\%, while reducing
irrelevant-evidence utilization to 10.6\% and 9.1\%, respectively. These
results demonstrate the complementary roles of the two rewards: EALR
encourages the policy to respond to retrieved information, whereas outcome
supervision favors evidence use that contributes to correct task completion,
resulting in more selective use of answer-supportive evidence. We further investigate whether CIPO promotes selective rather than indiscriminate evidence dependence through counterfactual masking experiments in Appendix E.

\begin{table}[h]
	\centering
	\small
	\setlength{\tabcolsep}{3.5pt}
	
	\begin{tabular}{cccccccc}
		\toprule
		Model & Outcome & EALR & Sup.$\uparrow$ & Irr.$\downarrow$ & ID & OOD & Avg. \\
		\midrule
		\multirow{3}{*}{3B}
		& \cmark &  & 18.6 & 12.4 & 0.326 & 0.349 & 0.336 \\
		&  & \cmark & 43.8 & 31.7 & 0.407 & 0.374 & 0.393 \\
		& \cmark & \cmark
		& \textbf{55.2} & \textbf{10.6}
		& \textbf{0.496} & \textbf{0.402} & \textbf{0.456} \\
		\midrule
		\multirow{3}{*}{7B}
		& \cmark &  & 21.9 & 11.3 & 0.358 & 0.392 & 0.372 \\
		&  & \cmark & 48.5 & 28.9 & 0.437 & 0.422 & 0.430 \\
		& \cmark & \cmark
		& \textbf{60.7} & \textbf{9.1}
		& \textbf{0.544} & \textbf{0.452} & \textbf{0.504} \\
		\bottomrule
	\end{tabular}
    \caption{Reward ablation summarized by domain-level F1. ID and OOD are
	averages over the four in-domain and three out-of-domain benchmarks. Full
	results are in Appendix A.}
	\label{tab:ablation}
\end{table}

\subsection{Analysis of Prior-Driven and Evidence-Driven}

We evaluate prior-driven reasoning on the test sets of all seven benchmarks using an LLM-assisted evaluation pipeline. All details are provided in Appendix F. Figure~\ref{fig:prior_driven_analysis} traces prior-driven rates during
training on all validation samples and on the subset of correctly answered
samples. In both views, IGPO's measured prior-driven rate increases over
training, whereas \method{} maintains a lower rate and decreases steadily.
The widening gap shows that \method{} more consistently discourages trajectories
whose answer-supporting evidence originates from the model's parametric
knowledge rather than from retrieved responses, including among trajectories
that reach the correct final answer.

\begin{figure}[t]
	\centering
	\includegraphics[width=0.49\columnwidth]
	{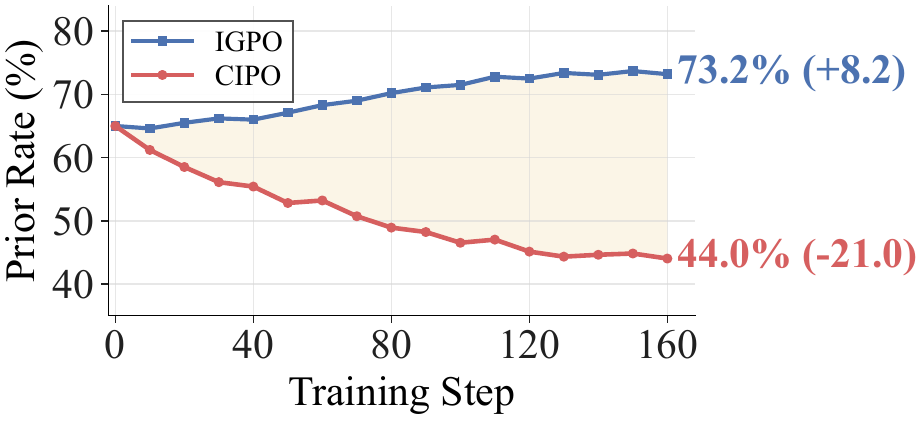}
	\hfill
	\includegraphics[width=0.49\columnwidth]
	{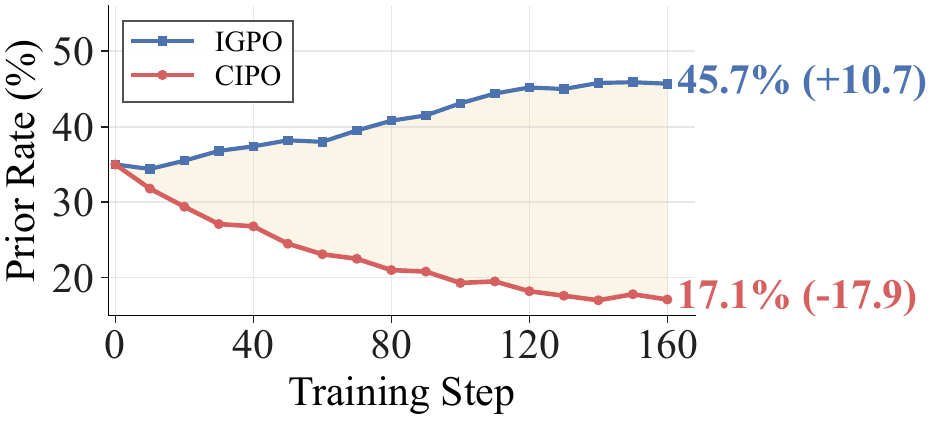}

	\makebox[0.49\columnwidth][c]{\small
	(a) All validation samples}
	\hfill
	\makebox[0.49\columnwidth][c]{\small
	(b) Correctly answered samples}

	\caption{Prior-driven rates during training. \method{} maintains lower
	measured rates than IGPO on both all validation samples and correctly
	answered samples.}
	\label{fig:prior_driven_analysis}
\end{figure}

\begin{figure}[t]
    \centering
    \includegraphics[width=0.49\columnwidth]
    {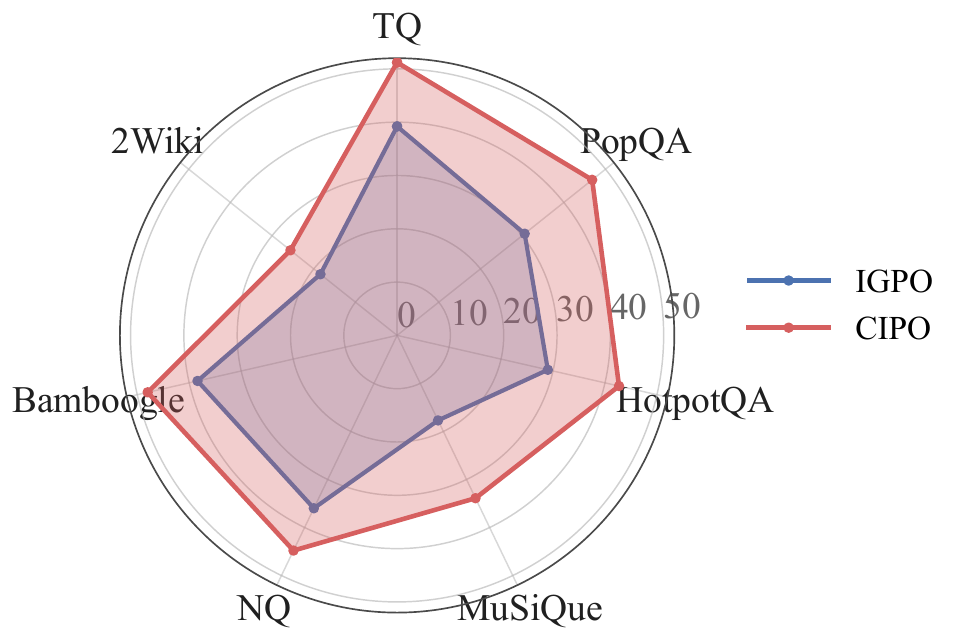}
    \hfill
    \includegraphics[width=0.49\columnwidth]
    {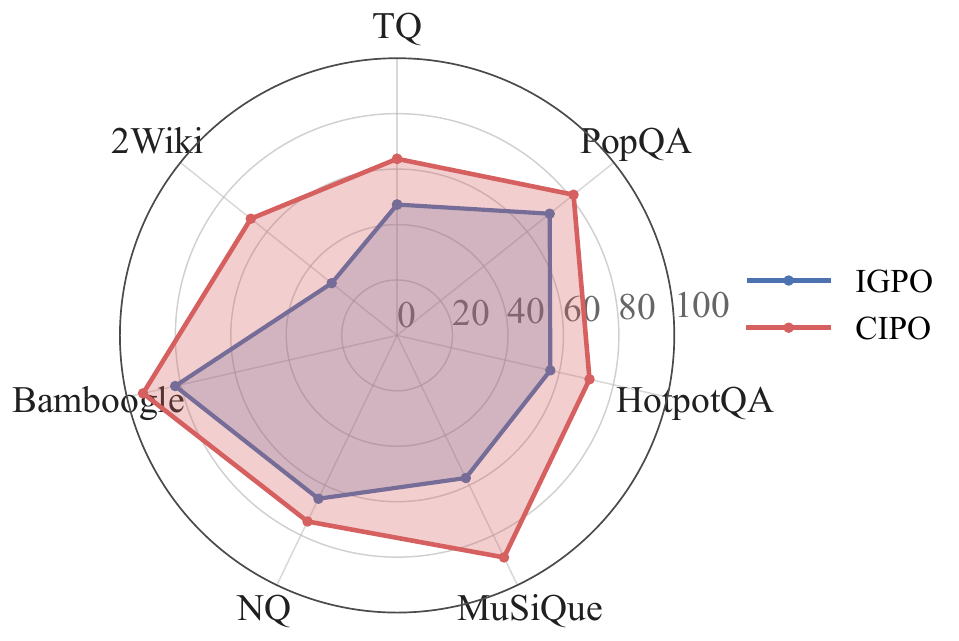}

    \makebox[0.49\columnwidth][c]{\small
    (a) All test samples}
    \hfill
    \makebox[0.49\columnwidth][c]{\small
    (b) Correctly answered samples}

    \caption{Evidence-driven rates on the test sets across seven benchmarks.
\method{} improves the measured rate on all test samples and on correctly
answered trajectories.}
    \label{fig:evidence_driven_analysis}
\end{figure}

Figure~\ref{fig:evidence_driven_analysis} provides the complementary
test-set analysis across the seven benchmarks. \method{} achieves higher
evidence-driven rates than IGPO on both all test samples and correctly
answered trajectories, yielding a larger radar area in both panels. In
particular, the overall evidence-driven rate increases from 29.9\% for IGPO
to 41.9\% for \method{}, while the rate among correct trajectories increases
from 59.4\% to 70.2\%. Under this diagnostic, these results indicate that CIPO more often grounds its
subsequent reasoning steps in evidence returned by the retrieval service.

\subsection{Comparison with RL Algorithms}

We compare \method{} with RLOO, PPO, GRPO, DAPO, and GSPO using
Qwen2.5-7B-Instruct as the base model. Table~\ref{tab:rl_algorithms} reports
representative single-hop and multi-hop results from both the in-domain and
out-of-domain evaluation regimes. Complete results on all seven benchmarks
are provided in Appendix~A. Among the evaluated approaches, \method{} achieves
the highest average F1 of 0.504, exceeding the strongest competing average,
achieved by GSPO (0.454), by 5.0 F1 points. The improvement is also reflected
across all four representative benchmarks: \method{} improves over the
strongest alternative from 0.422 to 0.467 on NQ, from 0.358 to 0.460 on
2Wiki, from 0.290 to 0.322 on MuSiQue, and from 0.501 to 0.552 on Bamboogle.
These gains cover both single-hop and multi-hop question answering in the two
evaluation regimes, demonstrating stronger aggregate performance under the
reported experimental configuration.
\begin{table}[t]
	\centering
	\small
	\setlength{\tabcolsep}{3.0pt}
	
	\begin{tabular*}{\columnwidth}{@{\extracolsep{\fill}}lccccc@{}}
		\toprule
		\multirow{2}{*}{Algorithm}
		& \multicolumn{2}{c}{In-domain}
		& \multicolumn{2}{c}{Out-of-domain}
		& \multirow{2}{*}{Avg.} \\
		\cmidrule(lr){2-3} \cmidrule(lr){4-5}
		& NQ & 2Wiki & MuSiQue & Bamboogle & \\
		\midrule
		RLOO & 0.378 & 0.323 & 0.227 & 0.457 & 0.410 \\
		PPO  & 0.394 & 0.322 & 0.233 & 0.489 & 0.421 \\
		GRPO & 0.411 & 0.346 & 0.255 & 0.476 & 0.429 \\
		DAPO & 0.401 & \underline{0.358} & 0.264 & \underline{0.501}
		     & 0.443 \\
		GSPO & \underline{0.422} & 0.353 & \underline{0.290} & 0.498
		     & \underline{0.454} \\
		\method{} & \textbf{0.467} & \textbf{0.460} & \textbf{0.322}
		& \textbf{0.552} & \textbf{0.504} \\
		\bottomrule
	\end{tabular*}
    \caption{Comparison with alternative RL algorithms on Qwen2.5-7B-Instruct. Full results are in Appendix A.}
	\label{tab:rl_algorithms}
\end{table}

\subsection{Training Dynamics Across Benchmarks}

Figure~\ref{fig:training_dynamics_main} tracks validation-set F1 over 80
training steps for \method{} and IGPO on two in-domain benchmarks, TriviaQA
and HotpotQA, and two out-of-domain benchmarks, MuSiQue and Bamboogle. Across
all four representative datasets, \method{} reaches higher F1 than IGPO early
in training and maintains this advantage at later training steps. Despite
dataset-specific fluctuations, this separation persists after the early
training phase and remains visible near the final evaluation step. The
trajectory-level comparison shows that the final performance gains of
\method{} are reflected throughout training in both evaluation regimes rather
than emerging only at the final checkpoint. Curves for the remaining benchmarks are provided in Appendix~B for a complete
benchmark-level view of training behavior across all evaluated settings.

\subsection{Efficiency Analysis}

Following the efficiency analysis in Search-R2~\cite{he2026searchr2}, we assess the practical training overhead of EALR relative to GRPO. EALR requires one additional evidence-masked scoring pass during training.
Table~\ref{tab:efficiency} reports the resulting per-step wall-clock time
relative to GRPO, together with the relative average-F1 improvement. The
G/C metric is defined as the relative F1 gain divided by the relative
training-time increase. For Qwen2.5-3B-Instruct, the per-step time increases
from 223.6 to 244.1 seconds, corresponding to a 9.17\% overhead, while the
relative average-F1 gain is 12.00\%. For Qwen2.5-7B-Instruct, the per-step
time increases from 417.2 to 436.5 seconds, or 4.63\%, together with a
17.48\% relative average-F1 gain. The gain-to-cost ratios are 1.31
and 3.78, respectively, indicating that the relative F1 improvements exceed
the corresponding relative training-time increases for both backbones. EALR
is used only to construct training rewards. At inference time, \method{}
requires neither evidence masking nor additional scoring passes beyond the
standard search-agent interaction loop.

\begin{table}[t]
	\centering
	\small
	\setlength{\tabcolsep}{3.0pt}
	
	\begin{tabular}{ccccc}
		\toprule
		Model & Time (s) & $\Delta$Time & $\Delta$F1 & G/C \\
		\midrule
		Qwen2.5-3B-Instruct & 223.6$\rightarrow$244.1 & 9.17\% & 12.00\% & 1.31 \\
		Qwen2.5-7B-Instruct & 417.2$\rightarrow$436.5 & 4.63\% & 17.48\% & 3.78 \\
		\bottomrule
	\end{tabular}
    \caption{Training efficiency of \method{} relative to GRPO.}
	\label{tab:efficiency}
\end{table}

\section{Conclusion}

In this work, we proposed \method{}, a reinforcement learning framework for
search agents that rewards evidence-sensitive reasoning while retaining
outcome supervision. CIPO assigns dense credit to actions whose generation
depends on retrieved evidence. Across seven benchmarks, it improves performance and reduces prior-driven reasoning. All results highlight the
value of explicitly rewarding evidence use.

In the future, we plan to expand CIPO to handle longer tasks with multiple tools and test it with different search tools, types of data and constantly changing information. We'll also work on better ways to make the training process for evidence-masked scoring less costly.

\bibliography{aaai2027} 

\onecolumn
\appendix

	\section{Full Experimental Results}
	\label{app:full_results}
	
	The main paper reports compact domain-level summaries for the reward ablation and RL-algorithm comparison. Tables~\ref{tab:ablation_full} and~\ref{tab:rl_algorithms_full} provide the complete per-dataset results.
	
	\begin{table}[!htbp]
		\centering
		\small
		\setlength{\tabcolsep}{5.0pt}
		\renewcommand{\arraystretch}{1.05}
		\caption{Full per-dataset ablation results for the reward design. Checkmarks indicate whether final-answer outcome reward and the EALR evidence-access sensitivity reward are used.}
		\label{tab:ablation_full}
		\resizebox{0.95\textwidth}{!}{%
			\begin{tabular}{cccccccccc}
				\toprule
				\multirow{2}{*}{Outcome}
				& \multirow{2}{*}{EALR}
				& \multicolumn{4}{c}{In-domain}
				& \multicolumn{3}{c}{Out-of-domain}
				& \multirow{2}{*}{Avg.} \\
				\cmidrule(lr){3-6} \cmidrule(lr){7-9}
				& & NQ & TQ & HotpotQA & 2Wiki & MuSiQue & Bamboogle & PopQA & \\
				\midrule
				\multicolumn{10}{l}{\textbf{Qwen2.5-3B-Instruct}} \\
				\cmark &  & 0.279 & 0.325 & 0.337 & 0.363 & 0.213 & 0.372 & 0.461 & 0.336 \\
				& \cmark & 0.365 & 0.482 & 0.384 & 0.395 & 0.256 & 0.401 & 0.465 & 0.393 \\
				\cmark & \cmark & \textbf{0.458} & \textbf{0.652} & \textbf{0.438} & \textbf{0.437} & \textbf{0.307} & \textbf{0.427} & \textbf{0.471} & \textbf{0.456} \\
				\midrule
				\multicolumn{10}{l}{\textbf{Qwen2.5-7B-Instruct}} \\
				\cmark &  & 0.285 & 0.362 & 0.401 & 0.382 & 0.223 & 0.481 & 0.472 & 0.372 \\
				& \cmark & 0.381 & 0.522 & 0.453 & 0.392 & 0.256 & 0.512 & 0.478 & 0.430 \\
				\cmark & \cmark & \textbf{0.467} & \textbf{0.726} & \textbf{0.521} & \textbf{0.460} & \textbf{0.322} & \textbf{0.552} & \textbf{0.482} & \textbf{0.504} \\
				\bottomrule
			\end{tabular}
		}
	\end{table}
	
	\begin{table}[!htbp]
		\centering
		\small
		\setlength{\tabcolsep}{5.0pt}
		\renewcommand{\arraystretch}{1.05}
		\caption{Full per-dataset comparison with alternative reinforcement learning algorithms on Qwen2.5-7B-Instruct. Bold and underlined values indicate the best and second-best results, respectively.}
		\label{tab:rl_algorithms_full}
		\resizebox{0.95\textwidth}{!}{%
			\begin{tabular}{lcccccccc}
				\toprule
				\multirow{2}{*}{Algorithm}
				& \multicolumn{4}{c}{In-domain}
				& \multicolumn{3}{c}{Out-of-domain}
				& \multirow{2}{*}{Avg.} \\
				\cmidrule(lr){2-5} \cmidrule(lr){6-8}
				& NQ & TQ & HotpotQA & 2Wiki & MuSiQue & Bamboogle & PopQA & \\
				\midrule
				RLOO & 0.378 & 0.655 & 0.443 & 0.323 & 0.227 & 0.457 & 0.384 & 0.410 \\
				PPO & 0.394 & 0.637 & 0.448 & 0.322 & 0.233 & 0.489 & 0.423 & 0.421 \\
				GRPO & 0.411 & 0.655 & 0.444 & 0.346 & 0.255 & 0.476 & 0.417 & 0.429 \\
				DAPO & 0.401 & \underline{0.686} & 0.462 & \underline{0.358} & 0.264 & \underline{0.501} & 0.431 & 0.443 \\
				GSPO & \underline{0.422} & 0.678 & \underline{0.481} & 0.353 & \underline{0.290} & 0.498 & \underline{0.456} & \underline{0.454} \\
				\method{} & \textbf{0.467} & \textbf{0.726} & \textbf{0.521} & \textbf{0.460} & \textbf{0.322} & \textbf{0.552} & \textbf{0.482} & \textbf{0.504} \\
				\bottomrule
			\end{tabular}
		}
	\end{table}
	
	\section{Additional Training Dynamics}
	\label{app:additional_training_dynamics}
	
\begin{figure}[!htbp]
		\centering
		\begin{minipage}[t]{0.32\textwidth}
			\centering
			\includegraphics[width=\linewidth]{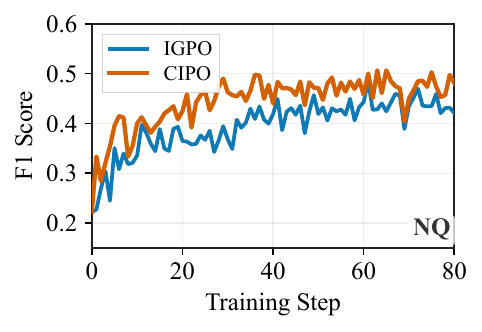}\\[-2pt]
			\footnotesize (a) Natural Questions (ID)
		\end{minipage}\hfill
		\begin{minipage}[t]{0.32\textwidth}
			\centering
			\includegraphics[width=\linewidth]{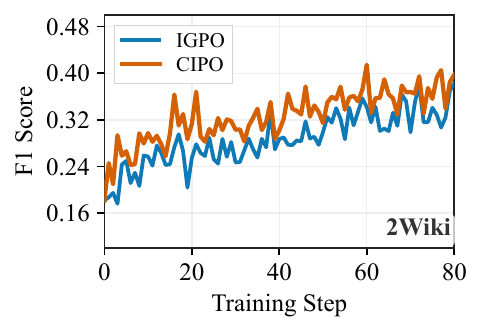}\\[-2pt]
			\footnotesize (b) 2WikiMultiHopQA (ID)
		\end{minipage}\hfill
		\begin{minipage}[t]{0.32\textwidth}
			\centering
			\includegraphics[width=\linewidth]{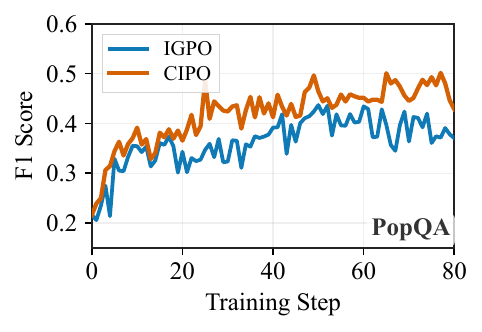}\\[-2pt]
			\footnotesize (c) PopQA (OOD)
		\end{minipage}
		
		\caption{Additional step-wise validation-set F1 trajectories for the benchmarks not shown in the main paper. Each panel compares \method{} with IGPO over 80 training steps and uses a dataset-specific vertical scale.}
		\label{fig:training_dynamics_additional}
	\end{figure}

	Figure~\ref{fig:training_dynamics_additional} reports the three benchmark trajectories omitted from the main paper's compact four-panel comparison.
	
\section{Hyperparameter Settings and Prompt Template}
    Table~\ref{tab:additional_hyperparameters} summarizes the important
optimization, rollout, and system configurations. The following prompt is used for training, validation, and test-time
rollouts. It is reproduced below verbatim to make the provenance of the
prompt and the tool-use interface explicit.

\begin{table}[t]
    \centering
    \caption{Additional training hyperparameters.}
    \label{tab:additional_hyperparameters}
    \renewcommand{\arraystretch}{1.05}
    \setlength{\tabcolsep}{5pt}
    \begin{tabular}{cc}
        \hline
        \textbf{Training Hyperparameters} & \textbf{Value} \\
        \hline
        Validation Batch Size
            & 256 \\
        Maximum Prompt Length
            & 30,767 \\
        Maximum Response Length
            & 2,000 \\
        Maximum Model Length
            & 32,768 \\
        Maximum Batched Tokens
            & 32,768 \\
        Actor Learning Rate
            & $1 \times 10^{-6}$ \\
        Critic Learning Rate
            & $1 \times 10^{-5}$ \\
        PPO Mini-Batch Size
            & 512 \\
        Actor PPO Micro-Batch / GPU
            & 2 \\
        Critic PPO Micro-Batch / GPU
            & 2 \\
        Rollout Log-Prob. Micro-Batch / GPU
            & 1 \\
        Reference Log-Prob. Micro-Batch / GPU
            & 2 \\
        Maximum PPO Tokens / GPU
            & 12,288 \\
        Rollout Temperature
            & 1.0 \\
        Actor KL Loss
            & Enabled \\
        KL-Loss Coefficient
            & 0.001 \\
        Actor Loss Aggregation
            & \texttt{token-mean} \\
        Tensor Parallel Size
            & 1 \\
        Ulysses Sequence Parallel Size
            & 4 \\
        vLLM GPU Memory Utilization
            & 0.55 \\
        Padding Removal
            & Enabled \\
        Dynamic Batch Size
            & Enabled \\

        \hline
    \end{tabular}
\end{table}

	\label{app:prompt_template}
	\begin{verbatim}
		* Today is {today}
		* You are an AI Assistant*
		
		The question I give you is a complex question that requires a *deep research*
		to answer. I will provide you with tools to help you answer the question:
		{%- for tool in tools.values() %}
		- {{ tool.name }}: {{ tool.description }}
		{%- endfor %}
		
		You don't have to answer the question now, but you should first think about
		the research plan or what to search next. Your output format should be one of
		the following two formats:
		
		# Rollout
		<think>
		YOUR THINKING PROCESS
		</think>
		<answer>
		YOUR ANSWER AFTER GETTING ENOUGH INFORMATION
		</answer>
		
		or
		
		<think>
		YOUR THINKING PROCESS
		</think>
		<tool_call>
		YOUR TOOL CALL WITH CORRECT FORMAT
		</tool_call>
		
		You should always follow the above two formats strictly. Only output the
		final answer (in words, numbers or phrase) inside the <answer></answer> tag,
		without any explanations or extra information. If this is a yes-or-no
		question, you should only answer yes or no.
		
		# Tools
		You may call one or more functions to assist with the user query.
		
		You are provided with function signatures within <tools></tools> XML tags:
		<tools>
		{%- for tool in tools.values() %}
		{{ '{' }}"type": "function", "function": {{ '{' }}"name": "{{ tool.name | replace("'", '"') }}", "description": "{{ tool.description }}", "parameters": {{ '{' }}"type": "object", "properties": {{tool.inputs | replace("'", '"')}}, "example": {{tool.example | replace("'", '"')}}, "uniqueItems": true{{ '}}}' }}
		{%- endfor %}
		</tools>
		
		For each function call, return a json object with function name and arguments
		within <tool_call></tool_call> XML tags:
		<tool_call>
		{{ '{' }}"name": <function-name>, "arguments": <args-json-object>{{ '}' }}
		</tool_call>
	\end{verbatim}

\section{Analysis of EALR Target Scope}
\label{app:ealr_scope}

EALR measures the influence of the most recently retrieved evidence
$E_{i,t}$ on the immediate next-turn action, rather than on the entire
remaining trajectory. This local formulation provides a clearer
turn-level attribution: by comparing the likelihood of the same
next-turn action under evidence-visible and evidence-masked conditions,
EALR directly evaluates whether the newly retrieved evidence affects the
agent's subsequent reasoning and search decision. In contrast, scoring
the full future trajectory would mix the effects of later reasoning
steps, newly retrieved evidence, and accumulated context, making it
difficult to attribute the likelihood difference specifically to
$E_{i,t}$.

Restricting EALR to the next turn does not ignore the long-term influence
of the current decision. CIPO separates local evidence sensitivity from
global task optimization: EALR provides dense rewards for individual
retrieval--decision transitions, while future intermediate rewards and
the terminal outcome reward are propagated through the turn-level
advantage
\begin{equation}
    \hat{A}_{i,t}
    =
    \sum_{j=t}^{T_i}
    \gamma^{j-t}\tilde{r}_{i,j}.
\end{equation}
This separation also ensures that EALR is computed only over
policy-generated \texttt{<think>} and \texttt{<search>} tokens, without
introducing additional environment-generated \texttt{<information>}
blocks from later turns.

\paragraph{Empirical Verification.}
We compare the proposed next-turn formulation with a full-trajectory
variant that computes the evidence-visible and evidence-masked
likelihood difference over all remaining model-generated actions. Both
variants use the same backbone, training data, retrieval environment,
and optimization settings.

\begin{table}[t]
    \centering
    \small
    \setlength{\tabcolsep}{5pt}
    \renewcommand{\arraystretch}{1.08}
    \caption{Comparison of different EALR target scopes on
    Qwen2.5-3B-Instruct. The proposed next-turn formulation consistently
    outperforms the full-trajectory variant.}
    \label{tab:ealr_scope}
    \resizebox{0.95\textwidth}{!}{%
        \begin{tabular}{lcccccccc}
            \toprule
            \multirow{2}{*}{EALR Target Scope}
            & \multicolumn{4}{c}{In-domain}
            & \multicolumn{3}{c}{Out-of-domain}
            & \multirow{2}{*}{Avg.} \\
            \cmidrule(lr){2-5} \cmidrule(lr){6-8}
            & NQ & TQ & HotpotQA & 2Wiki
            & MuSiQue & Bamboogle & PopQA & \\
            \midrule
            Full future trajectory
            & 0.315 & 0.512 & 0.302 & 0.395
            & 0.258 & 0.288 & 0.321 & 0.342 \\

            Next-turn action (Ours)
            & \textbf{0.458} & \textbf{0.652}
            & \textbf{0.438} & \textbf{0.437}
            & \textbf{0.307} & \textbf{0.427}
            & \textbf{0.471} & \textbf{0.456} \\
            \bottomrule
        \end{tabular}%
    }
\end{table}

The next-turn formulation outperforms the full-trajectory variant on
all seven benchmarks, improving the average F1 from 0.342 to 0.456.
The substantial gain of 11.4 F1 points indicates that extending the
scoring target to the full trajectory introduces interference from
later decisions and retrieval results, thereby weakening turn-level
credit assignment. These results support using the immediate next-turn
action as the target of EALR.

\section{Selective Evidence Dependence under Counterfactual Masking}
\label{app:selective_evidence_dependence}
Before presenting the counterfactual analysis, we define the evidence utilization metrics used throughout this section.
Supportive evidence utilization (Sup.) measures the proportion of trajectories whose reasoning actions are judged to depend on retrieved evidence that directly supports the reference answer.
Irrelevant evidence utilization (Irr.) measures the proportion of trajectories whose reasoning actions rely on retrieved evidence that does not contribute to solving the question or contradicts the required reasoning path.
The evidence annotations are obtained using the same evidence identification procedure described in Appendix F. Higher Sup. and lower Irr. indicate more selective and task-relevant evidence dependence.

Although EALR measures the sensitivity of policy actions to accessible
evidence, an important question is whether CIPO merely encourages
indiscriminate responsiveness to retrieved text. To examine this issue,
we construct three counterfactual masking variants. \emph{Random-context
Mask} replaces the evidence mask with an equally sized randomly selected
context span, testing whether the improvement can be attributed to
generic context perturbation. \emph{Supportive-Evidence-Only Mask}
computes the sensitivity signal using only answer-supportive evidence
spans, serving as an oracle-style variant that requires evidence-level
annotations. \emph{Non-supportive-Evidence-Only Mask} instead applies
the mask only to irrelevant or non-supportive evidence. All variants
retain the same terminal outcome reward and differ only in the context
span used to construct the EALR reward.

We evaluate the resulting policies using supportive-evidence utilization
(Sup.), irrelevant-evidence utilization (Irr.), Evidence Selectivity, and
question-answering performance. Evidence Selectivity measures whether the
policy depends more strongly on answer-supportive evidence than on
non-supportive evidence, with a higher value indicating more selective
and task-relevant evidence dependence.

\begin{table}[t]
    \centering
    \small
    \setlength{\tabcolsep}{6pt}
    \renewcommand{\arraystretch}{1.08}
    \caption{Counterfactual masking analysis on
    Qwen2.5-3B-Instruct. All variants use the same terminal outcome
    reward and differ only in the span masked when constructing the
    evidence-sensitivity reward.}
    \label{tab:counterfactual_masking}
    \resizebox{0.95\textwidth}{!}{%
        \begin{tabular}{lcccccc}
            \toprule
            Method
            & Sup.$\uparrow$
            & Irr.$\downarrow$
            & Evidence Selectivity$\uparrow$
            & ID
            & OOD
            & Avg.\ F1 \\
            \midrule
            Random-context Mask
            & 35.0
            & 34.5
            & 2.1
            & 0.415
            & 0.338
            & 0.382 \\

            Supportive-Evidence-Only Mask
            & \textbf{62.5}
            & \underline{14.2}
            & \underline{60.2}
            & \underline{0.488}
            & \underline{0.395}
            & \underline{0.448} \\

            Non-supportive-Evidence-Only Mask
            & 28.4
            & 45.3
            & -20.5
            & 0.375
            & 0.305
            & 0.345 \\

            \method{} (Ours)
            & \underline{55.2}
            & \textbf{10.6}
            & \textbf{66.7}
            & \textbf{0.496}
            & \textbf{0.402}
            & \textbf{0.456} \\
            \bottomrule
        \end{tabular}%
    }
\end{table}

As shown in Table~\ref{tab:counterfactual_masking}, randomly masking an
equally sized context span yields an Evidence Selectivity score of only
2.1, despite obtaining moderate supportive- and irrelevant-evidence
utilization rates. This result indicates that generic masking or context
perturbation does not naturally produce selective evidence use.
Moreover, masking only non-supportive evidence leads to negative
selectivity ($-20.5$), the highest irrelevant-evidence utilization
(45.3), and the lowest average F1 (0.345). Thus, encouraging sensitivity
to arbitrary retrieved content can reinforce dependence on unhelpful
evidence and degrade task performance.

The supportive-evidence-only variant substantially improves evidence
utilization and reaches an average F1 of 0.448, confirming that assigning
credit to task-relevant evidence is beneficial. However, this variant
requires explicit identification of answer-supportive evidence. In
contrast, CIPO operates on the complete retrieved evidence block without
evidence-level supervision, yet achieves the lowest irrelevant-evidence
utilization (10.6), the highest Evidence Selectivity (66.7), and the best
average F1 (0.456). Although its supportive-evidence utilization is
slightly lower than that of the oracle-style supportive-only variant,
CIPO more effectively suppresses non-supportive evidence, resulting in
stronger overall selectivity and task performance.

These results show that CIPO does not simply increase the policy's
sensitivity to all retrieved text. Instead, combining evidence-access
sensitivity with outcome supervision encourages the policy to retain
dependencies that contribute to correct task completion while
suppressing dependencies on irrelevant evidence. CIPO therefore promotes
selective and task-relevant evidence dependence rather than
indiscriminate evidence responsiveness.

\section{LLM-assisted Evaluation Protocol}
\label{app:llm_evaluation_protocol}

This appendix provides the detailed evaluation protocol for measuring
prior-driven and evidence-driven reasoning behaviors described in
Section~4.4. We use an LLM-assisted pipeline to classify whether an
agent trajectory primarily relies on retrieved evidence or the model's
parametric knowledge.

\subsection{Evaluation Pipeline}

For each evaluated trajectory, we first identify evidence passages that
support the reference answer. For datasets with existing evidence
annotations, we directly use the provided annotations. For datasets without explicit supporting evidence annotations, we use
DeepSeek-V4-Flash to extract evidence passages that support the reference
answer. Each automatically identified evidence span is manually checked
to remove incorrect or insufficient evidence annotations. The verified evidence annotations are used only for evaluation and are
never provided as supervision during CIPO training. Therefore, the
behavioral evaluation does not introduce additional evidence-level
training signals.

Given the retrieved evidence, reasoning trajectory, and final answer, we
use GPT-5.5 as the external LLM judge to classify each trajectory as
\emph{prior-driven} or \emph{evidence-driven}. The judge determines
whether the key reasoning steps and final conclusion are primarily
supported by the retrieved evidence or are generated mainly from the
model's parametric knowledge. The same evaluation procedure is applied to both training-time
validation samples and final test samples.

\subsection{GPT-5.5 Judge Criteria}

The GPT-5.5 judge receives the question, retrieved evidence,
reasoning trajectory, and final answer. It is instructed to determine
whether the trajectory exhibits prior-driven or evidence-driven
reasoning according to the following criteria:

\begin{itemize}
    \item \textbf{Evidence-driven reasoning:}
    The agent's intermediate reasoning steps and final conclusion are
    substantially influenced by retrieved evidence. The retrieved
    information is used to verify, revise, or construct the answer.

    \item \textbf{Prior-driven reasoning:}
    The agent reaches the conclusion mainly from its internal
    parametric knowledge, while retrieved evidence only confirms an
    existing hypothesis or has little influence on subsequent reasoning.
\end{itemize}

The complete judging prompt used by GPT-5.5 is provided below.

\label{app:gpt55_prompt}

\begin{verbatim}
    [GPT-5.5 Judge Prompt]
    
    You are an expert evaluator for analyzing reasoning trajectories
    of search agents.

    Given:
    - Question:
    {question}

    - Retrieved Evidence:
    {evidence}

    - Agent Trajectory:
    {trajectory}

    - Final Answer:
    {answer}

    Determine whether the trajectory is primarily:

    (A) Evidence-driven:
    The reasoning process substantially uses retrieved evidence,
    and the evidence affects intermediate reasoning or the final answer.

    (B) Prior-driven:
    The reasoning process mainly relies on internal knowledge,
    while retrieved evidence only confirms or weakly affects
    the conclusion.

    Output:
    <classification>
    Evidence-driven or Prior-driven
    </classification>

    <explanation>
    Briefly explain the decision.
    </explanation>
\end{verbatim}

\end{document}